\documentclass{article}
\usepackage{microtype}
\usepackage{graphicx}
\usepackage{subfigure}
\usepackage{float}
\usepackage{booktabs} % for professional tables
\usepackage{multirow}
\usepackage{hyperref}

\usepackage[accepted]{icml2024}
\usepackage{amsmath}
\usepackage{amssymb}
\usepackage{mathtools}
\usepackage{amsthm}
\usepackage{balance}
\usepackage{flushend}
\usepackage{makecell}

\usepackage[capitalize,noabbrev]{cleveref}

\theoremstyle{plain}

\theoremstyle{definition}

\theoremstyle{remark}

\usepackage[textsize=tiny]{todonotes}
\usepackage{color, colortbl}
\usepackage{enumitem}
\usepackage{arydshln}

\setlist[itemize]{align=parleft,left=0pt..0.8em}
\definecolor{airforceblue}{rgb}{0.08, 0.38, 0.74}
\definecolor{blue1}{rgb}{0.796,0.878,0.937}
\begin{document}
	%	\SetKwComment{Comment}{/* }{ */}
	\newcommand{\sysname}{{LongRoPE}}
	\newcommand{\lz}[1]{{\textcolor{red}{\it Lyna: #1}}}	
	\twocolumn[
	%	\icmltitle{Beyond 2048K Tokens: Extending LLM Context Window  with  Non-Uniform Positional Interpolation and Extrapolation}
	\icmltitle{{\sysname}: Extending LLM Context Window Beyond 2 Million Tokens}
	
	%	\icmltitle{LLMs with 256K Tokens: Push the Limits of Position Interpolation for Context Window Extension with Searchable Non-uniformity}

	% It is OKAY to include author information, even for blind
	% submissions: the style file will automatically remove it for you
	% unless you've provided the [accepted] option to the icml2023
	% package.
	
	% List of affiliations: The first argument should be a (short)
	% identifier you will use later to specify author affiliations
	% Academic affiliations should list Department, University, City, Region, Country
	% Industry affiliations should list Company, City, Region, Country
	
	% You can specify symbols, otherwise they are numbered in order.
	% Ideally, you should not use this facility. Affiliations will be numbered
	% in order of appearance and this is the preferred way.
	\icmlsetsymbol{equal}{*}
\icmlsetsymbol{corres}{$\dagger$}
	\begin{icmlauthorlist}
			\vspace{-2ex}
		\icmlauthor{Yiran Ding}{equal}
		\icmlauthor{Li Lyna Zhang}{corres}
		\icmlauthor{Chengruidong Zhang}{}
		\icmlauthor{Yuanyuan Xu}{equal}
		\icmlauthor{Ning Shang}{}
		\icmlauthor{Jiahang Xu}{}\\
		\icmlauthor{Fan Yang}{}
		\icmlauthor{Mao Yang}{}\\
		%\icmlauthor{Yiran Ding}{ms,equal}
		%\icmlauthor{Li Lyna Zhang}{ms,corres}
	%	\icmlauthor{Chengruidong Zhang}{ms}
	%	\icmlauthor{Yuanyuan Xu}{ms,equal}
	%	\icmlauthor{Ning Shang}{ms}
		%\icmlauthor{Jiahang Xu}{ms}
	%	\icmlauthor{Fan Yang}{ms}
	%	\icmlauthor{Mao Yang}{ms}\\
	\vspace{1ex}
		\icmlauthor{\fontsize{9.5}{9.5} \selectfont{{Microsoft Research}}}{}
\end{icmlauthorlist}
	
%	\icmlaffiliation{ms}{Microsoft Research}
	
	\icmlcorrespondingauthor{Li Lyna Zhang}{lzhani@microsoft.com}
	
	\vskip 0.3in
	]
	\printAffiliationsAndNotice{\icmlEqualContribution} 
	
	%\twocolumn[
	%\icmltitle{{\sysname}: Scaling Up LLM Context Window  via Searchable Non-Uniform Position Interpolation and Extrapolation}
	%\vskip 0.3in
	%]
	%\vspace{-2ex}
\begin{abstract}
%	\vspace{-2ex}
Large context window is a desirable feature in large language models (LLMs). 
However, due to high fine-tuning costs, scarcity of long texts, and catastrophic values introduced by new token positions, current extended context windows are limited to around 128k tokens.

This paper introduces {\sysname} that,  for the first time, extends the context window of pre-trained LLMs to an impressive  \textbf{2048k} tokens,  with up to only 1k fine-tuning steps at within 256k training lengths, while maintaining performance at the original short context window. This is achieved 
by three key innovations: (i) we identify and exploit two forms of non-uniformities in positional interpolation through an efficient  search, providing a better initialization for fine-tuning and  enabling an 8$\times$ extension in non-fine-tuning scenarios; (ii) we introduce a progressive extension strategy that first fine-tunes a 256k length LLM and then conducts a second  positional interpolation on the fine-tuned extended LLM to achieve a 2048k context window; (iii) we readjust {\sysname} on 8k length to recover the short context window performance. 
Extensive experiments on LLaMA2 and Mistral across various tasks demonstrate the effectiveness of our method.  Models extended via {\sysname} retain the original architecture with minor modifications to the positional embedding,  and can reuse most pre-existing optimizations. Code will be available at \url{https://github.com/microsoft/LongRoPE}

\end{abstract}
	%\vspace{-5ex}
\section{Introduction}
%\vspace{-1ex}
Large Language Models (LLMs), despite remarkable success on various tasks~\cite{gpt4,llama2}, often suffer from limited context window size, e.g., LLaMA2's 4096 token limit~\cite{llama2}. 
Beyond the context window, LLM's performance declines due to the additional positions that the model has not been trained on. This poses challenges in important scenarios like in-context learning with numerous examples~\cite{Huang2023FewerIM} and LLM agents~\cite{park2023generative,madaan2023selfrefine}. 

Recent works show that a pre-trained LLM context window can be extended to around 128k by fine-tuning on longer texts~\cite{longlora,pi,yarn,zhang2024soaring,liu2023scaling}.  
There are three major obstacles to further extend the context window.
First, 
untrained new position indices introduce many catastrophic values, leading to out-of-distribution issues and  making fine-tuning difficult to converge~\cite{pi}. This is particularly challenging when an extension from 4k to  $>$1000k introduces more than 90\% new positions. Second, fine-tuning usually requires texts of corresponding lengths. However, long texts in current datasets, especially those exceeding 1000k, are limited. Moreover, training on extra-long texts is computationally expensive, requiring prohibitively extensive training hours and GPU resources. Third, when extending to extremely long context windows, the attention becomes dispersed as it's spread thinly across numerous token positions, degrading performance on the original short context~\cite{pi}.

 \begin{figure}[t]
	\centering
	\includegraphics[width=1\columnwidth]{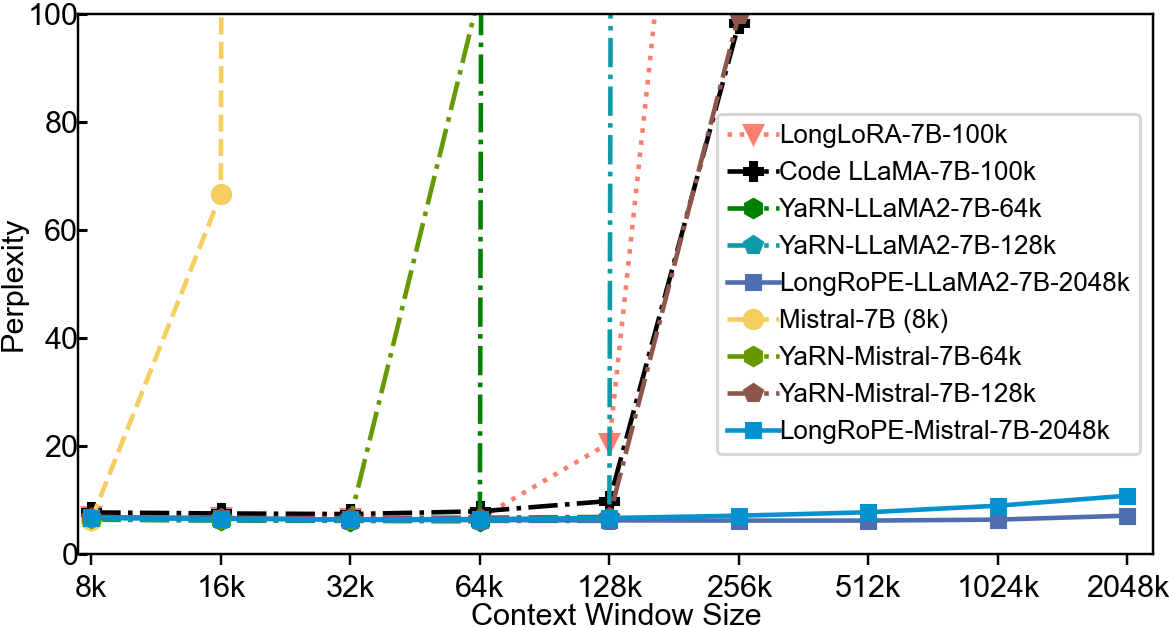}
	\vspace{-5ex}
	\caption{Books3 perplexity comparison between {\sysname} and state-of-the-art long-context LLMs using other extension methods. }
	\label{fig:finalppl}
\end{figure}

\begin{figure*}[ht]
	\centering
	\includegraphics[width=1\textwidth]{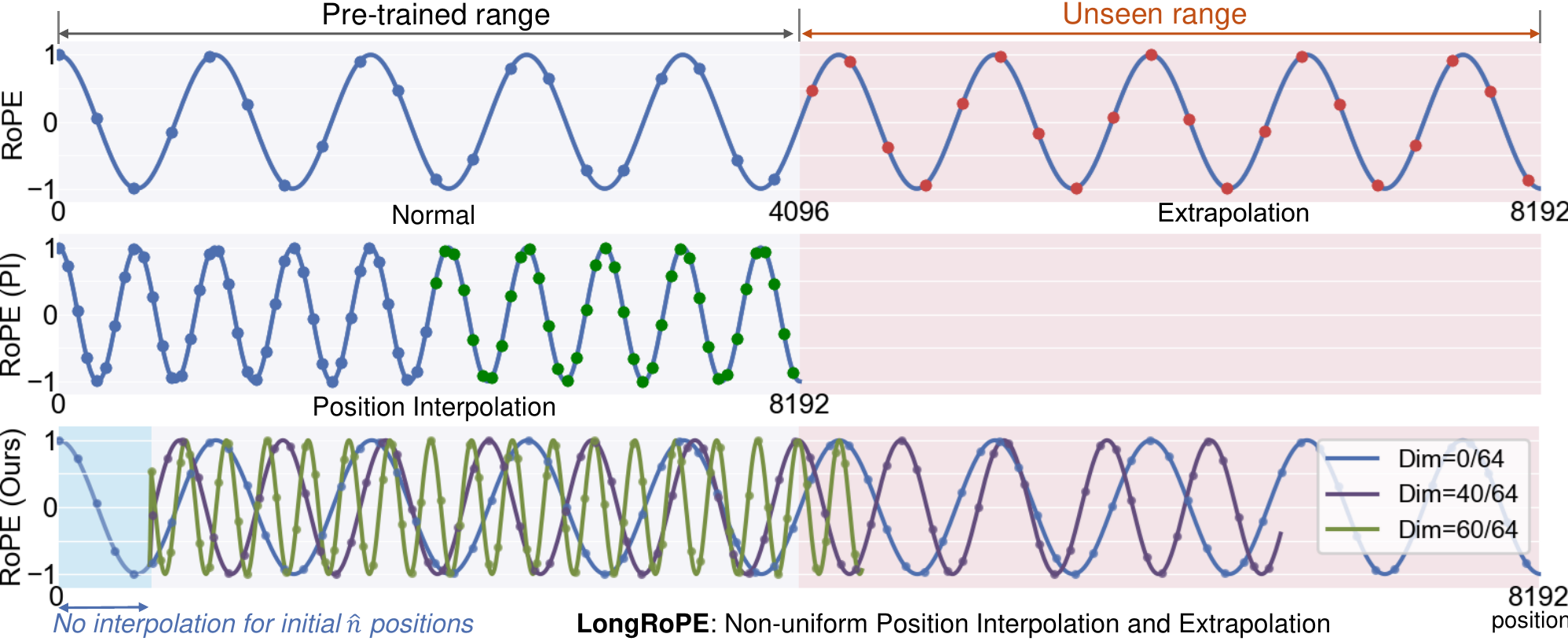}
	\vspace{-5ex}
	\caption{An  illustrative example to show RoPE embedding under different interpolation methods. \textit{Upper}: RoPE under direct extrapolation. \textit{Middle}: Rescaled RoPE under linear positional interpolation. \textit{Down}: {\sysname} fully exploits the identified two non-uniformities, leading to varied interpolation and extrapolation across RoPE dimensions at different token positions. }
	\label{fig:overview}
	\vspace{-2ex}
\end{figure*}

One approach to mitigate the first challenge is to interpolate RoPE positional embedding~\cite{rope,pi},  which downscales new position indices to the pretrained range, as shown in Fig.\ref{fig:overview}. Position Interpolation (PI) ~\cite{pi} linearly interpolates RoPE's rotary angles by the extension ratio. NTK~\cite{ntk,dynamicntk} advocates unequal interpolation and extrapolation across RoPE dimensions. YaRN~\cite{yarn} categorizes RoPE dimensions into three frequency-based groups and applies extrapolation, NTK, and linear interpolations, respectively. However,  positional embedding exhibits \emph{complex non-uniform information entropy} in the Transformer architecture. %Yet, existing approaches fail to effectively manage 
Such subtle non-uniformity is not effectively leveraged by existing approaches, leading to information loss and hence limiting the context window size.

Section~\ref{sec:analysis} reveals two key findings empirically: \textbf{(1)} Effective positional interpolation should consider two forms of non-uniformities: varying RoPE dimensions and token positions. Lower RoPE dimensions and initial starting token positions benefit from less interpolation, but the optimal solutions depend on the target extended length.
  \textbf{(2)} By considering these non-uniformities into positional interpolation, we can effectively retain information in the original RoPE,  particularly key dimensions and token positions. This minimizes the loss caused by positional interpolation, and thus provides better initialization for fine-tuning. Moreover, it  allows an 8$\times$ extension in non-fine-tuning scenarios.

  Motivated by the findings, we introduce \textbf{\sysname},  an effective method 
  that extends the LLM context window beyond 2 \emph{million} tokens.  {\sysname} is based on three key innovations. First, \sysname{} fully exploits multidimensional non-uniformities in positional interpolation. 
  It identifies effective rescale factors for RoPE's rotation angles for each RoPE dimension, based on token positions. 
  As the search space that identifies rescale factors expands exponentially with the target extension ratio, \sysname{} introduces an evolutionary search algorithm with two optimization techniques to boost search efficiency. Fig.~\ref{fig:overview} shows an example of the searched  rescaled RoPE.

 Then, \sysname{} leverages an efficient, progressive extension strategy to achieve a 2048k context window without the need of direct fine-tuning on texts with extremely long lengths, which are scarce and hardly available. The strategy begins by searching for a 256k length on the pre-trained LLM and fine-tuning it under this length. Then, as our non-uniform positional interpolation allows for an 8$\times$ extension in non-fine-tuning settings, we conduct a second search for new RoPE rescale factors on the fine-tuned extended LLM. This ultimately achieves the 2048k context window for LLaMA2 and Mistral~\cite{mistral}.
 
Finally, 
to mitigate performance degradation on the original (shorter) context window,
\sysname{} continues to adjust 
the RoPE rescale factors on the extended LLM. 
Similar to scaling up from 256k to 2048k, we scale down to 4k and 8k context windows on the 256k fine-tuned LLM using our search algorithm to encourage less positional interpolation. During inference, if the sequence length is less than 8k, we update RoPE with the searched rescale factors.

Extensive experiments across different LLMs and various long-context tasks demonstrate the effectiveness of our method. We show that {\sysname} is highly effective in maintaining low perplexity from 4k to 2048k evaluation length,  achieving over 90\% passkey retrieval accuracy, and delivering  comparable accuracy on standard benchmarks designed within the 4096 context window.  {\sysname} can be applied to any LLMs based on RoPE embedding. We will release our code and {\sysname}-2048k models.

	%\vspace{-2ex}
\section{Non-uniformity in Positional Interpolation}
\label{sec:analysis}
%\vspace{-1ex}
\subsection{Preliminary}
%\vspace{-1ex}
Transformer models require explicit positional information, often in the form of position embedding, to represent the order of input tokens. Our work focuses on the RoPE~\cite{rope} position embedding, which is widely used in recent  LLMs. For a token at position index $n$, its corresponding RoPE encoding can be simplified as follows:
\vspace{-1ex}
\begin{equation}
		\fontsize{7.8}{7.8} \selectfont
	\label{eq:rope}
	\left[ cos(n\theta_0), sin(n\theta_0), cos(n\theta_1), \cdot\cdot\cdot, cos(n\theta_{d/2-1}), sin(n\theta_{d/2-1}) \right]   
\end{equation}
where $d$ is the embedding dimension, 
 $n\theta_i$ is the rotary angle of token at position $n$, 
$\theta_i=\theta^{-2i/d}$ represents the rotation frequencies. In RoPE, the default base value of $\theta$ is 10000.

\noindent\textbf{\emph{Context window extension ratio $s$} and positional interpolation}. We define $s$ as  the ratio of extended context length $L'$ to the original length $L$: $s=\frac{L'}{L}$. 

To extend the context window from $L$ to $L'$, current positional interpolation methods
suggest downscaling rotation frequencies $\theta_i$ based on extension ratio $s$. Let $\beta=\theta^{2/d}$, and $\lambda$ denote the actual rescale factor related to $s$,  
we   unify these positional interpolation methods as follows:
\vspace{-1ex}
\begin{equation}	
	\fontsize{7.5}{7.5} \selectfont
	\label{eq:unified_rope}
	\begin{aligned}
		\left[cos\left(\frac{n}{\lambda(\beta)^0}\right), sin \left(\frac{n}{\lambda(\beta)^0}\right), cos\left(\frac{n}{\lambda(\beta)^1}\right), \cdot\cdot\cdot,  sin \left(\frac{n}{\lambda(\beta)^{d/2-1}}\right) \right]   
	\end{aligned}
\end{equation}

\noindent\textbf{Linear positional interpolation (PI)}. PI~\cite{pi}  suggests linear interpolation of position indices within the pre-trained length limit. 
For a target extension ratio $s$, the rotation angles of all positions are linearly reduced by $\lambda=s$  across all RoPE dimensions.
 However, this makes the position information very ``crowded", hindering the model's ability distinguish closely positioned tokens. Therefore, PI tends to underperform at high extension ratios.

\noindent\textbf{NTK-based interpolation and extrapolation}. 
~\cite{ntk,dynamicntk} look at RoPE from an information encoding perspective  and 
apply the Neural Tangent Kernel (NTK) theory~\cite{jacot2018neural,tancik2020fourier}.
To mitigate the crowded-positions issue in PI, they suggest to distribute 
 interpolation pressure across RoPE dimensions. 
 It scales lower (high frequency) dimensions less and higher (low frequency) dimensions more, resulting in  both positional interpolation and extrapolation, where $\lambda=s^{i}$. The improved dynamic NTK~\cite{dynamicntk}  adjusts the extension ratio at each position based on the current sequence length. Unlike PI, which necessitates fine-tuning, NTK-aware methods can extend context windows in non-fine-tuning scenarios, but usually with a maximum extension ratio of 4$\times$.

\noindent\textbf{YaRN}~\cite{yarn} introduces a significant improvement to positional interpolation performance. It divides RoPE dimensions into three frequency-based groups, each with a different interpolation strategy. High frequency dimensions undergo extrapolation ($\lambda$=1), while low frequency dimensions use linear interpolation (PI). The RoPE dimensions that fall in-between employs the NTK.
The key of YaRN lies in its grouping of RoPE dimensions, which currently depends on human-led empirical experiments. This may result in sub-optimal performance for new LLMs.

 %\vspace{-2ex}
\subsection{Study on Non-uniform Positional Interpolation}
%\vspace{-1ex}
%Our work aims to probe the boundaries of extending LLM context window via manipulating RoPE embedding. Despite these significant efforts, we identify several untapped optimization opportunities. Our intuition is that \textit{the current positional interpolation methods, which are heavily reliant on human heuristics and empirical rules,  may lead to sub-optimal performance}.
\begin{table}[t]
	\caption{Perplexity of LLaMA2-7B extended via different methods. By a simple search for the rescale factors of each RoPE dimension, we can greatly reduce the perplexity. }
	\label{tbl:exp1}
	\small
	\centering
	\resizebox{0.48\textwidth}{!}{
		\begin{tabular}
			{@{\hskip0pt}c|cc|c@{\hskip2pt}c@{\hskip0pt}}
			\toprule
			(LLaMA2-7B) &\multicolumn{4}{c@{\hskip0pt}}{Context Window Size}\\
					\cline{2-5}
			\multirow{2}{*}{Extension method}	&	\multicolumn{2}{c|}{PG19 (5 samples)}&\multicolumn{2}{c@{\hskip0pt}}{Proof-pile (10 samples)}\\
			\cline{2-5}
		& 8192&16384 & 8192 & 16384\\
			\hline
			PI&10.65&20.49&3.65&4.93\\
			Dy-NTK&10.21&23.29&3.50&3.87\\
			YaRN & 32.64& 87.89& 3.49& 3.25\\
			\hline
			\bf {Search for RoPE Dim-wise $\lambda$} & \bf9.37&\bf11.34&\bf3.45&\bf3.13\\
			\hline
	\end{tabular}}
\end{table}

\begin{table}[t]
	\caption{Perplexity of LLaMA2-7B extended on PG19 (5 samples). When retaining the first $\hat{n}$ tokens without positional interpolation, the performance of both PI and Dynamic-NTK are improved. }
	\label{tbl:tokenposition}
	\small
	\centering 
	\resizebox{0.48\textwidth}{!}{	\begin{tabular}{@{\hskip0pt}c@{\hskip2pt}|@{\hskip2pt}c@{\hskip2pt}|@{\hskip3pt}c@{\hskip3pt}c@{\hskip3pt}c@{\hskip3pt}c@{\hskip3pt}c@{\hskip3pt}c@{\hskip3pt}c@{\hskip3pt}c@{\hskip3pt}c@{\hskip0pt}}
			\toprule 
			(LLaMA2-7B)	&	\multirow{2}{*}{$L'$}&\multicolumn{9}{c}{No interpolation for first $\hat n$ tokens}\\
			\cline{3-11}
		Extension method&	&0&2&4&8&16&32&64&128&256\\
			\hline
			\multirow{2}{*}{PI} & 8k & 10.65&10.65&10.65&10.65&10.66&10.59&\bf 10.49&10.54&11.14\\
			&16k&20.49&20.39&20.36&20.02&\bf 19.64&19.96&20.64&22.27&34.65\\
			\hline 
			\multirow{2}{*}{Dy-NTK}&8k& 10.21&10.21&10.21&10.21&10.22&10.20&10.15&\bf 10.13&10.62\\
			&16k&23.29&23.29&23.27&23.08&\bf 22.68&22.94&23.58&27.94&90.99\\
			\hline
	\end{tabular}}
\end{table}

\begin{table}[t]
	\caption{Proof-pile perplexity of the extended LLaMA2-7B with a 64k context window in non-fine-tuned and fine-tuned settings.}
	\label{tbl:finetune}
	\small
	\centering 
		\begin{tabular}
			{c|cc}
			\toprule
			Method	&non-fine-tuned&fine-tuned\\
			\hline
			PI& 72.54& 2.44\\
			\hline
			YaRN & 4.15&2.42 \\
			\hline
			\textbf{Search (Dim-wise $\lambda$ and $\hat n$)} &\bf 3.22 &\bf 2.36 \\
			\hline		\end{tabular}%}
\end{table}

Inspired by NTK and YaRN, we notice their gains from non-linearity, specifically in considering different frequencies across RoPE dimensions for specialized interpolation and extrapolation.
However, current non-linearities heavily rely on human-designed rules. 
This naturally raises two questions: (1) Is the current positional interpolation optimal? (2) Are there unexplored  non-linearities? 

To answer these questions, we use evolution search (see Sec.~\ref{sec:method}) to discover better non-uniform positional interpolations for LLaMA2-7B. The search is guided by perplexity, using 5 random samples from PG19~\cite{pg19} validation set.  Through our empirical analysis, we reveal the following key findings.

\noindent\textbf{Finding 1}: \textit{RoPE dimensions exhibit substantial non-uniformities, which are not effectively handled by current positional interpolation methods.}

We  search the optimal $\lambda$ for each RoPE dimension in Eq.~\ref{eq:unified_rope}. Table~\ref{tbl:exp1} compares the perplexity of LLaMA2-7B under different  methods on PG19 and Proof-pile~\cite{proof-pile} test sets, without fine-tuning. Our searched solution shows significant improvements, suggesting that current linear (PI) and non-uniform (Dynamic-NTK and YaRN) interpolations are sub-optimal. Notably, YaRN underperforms  than PI and NTK on PG19, as it doesn't reach the target context window length for non-fine-tuned LLM. For example, YaRN's perplexity spikes after 7k in an 8k context size. 

Through our search,  the rescaled factors $\lambda$ in Eq.~\ref{eq:unified_rope} become non-uniform, differing from the fixed scale $s$ in PI, NTK's formula calculation, and YaRN's group-wise calculation. These non-uniform factors significantly improve LLaMA2's language modeling performance (i.e., perplexity) for 8k and 16k context windows without fine-tuning. This is because the resulting positional embedding effectively preserves the original RoPE, especially key dimensions, thus reducing LLM's difficulty in distinguishing close token positions.

\noindent\textbf{Finding 2}: \textit{RoPE for the initial tokens in the input sequence should be extrapolated with less interpolation.} 

For the initial $\hat{n}$ tokens in input sequences, we hypothesize that their RoPE should do less interpolation. This is because they receive large attention scores, making them crucial to attention layers, as observed in  Streaming LLM~\cite{streamingllm} and LM-Infinite~\cite{han2023lminfinite}. To verify this, 
 we extend the context window to 8k and 16k using PI and NTK, keeping the first  $\hat n$ (0,2, ..., 256) tokens without interpolation. When $\hat n$=0, it reverts to the original PI and NTK. Table~\ref{tbl:tokenposition} highlights two key observations: \textbf{(1)} retaining the starting tokens without position interpolation indeed improves the performance. \textbf{(2)} The optimal number of starting tokens, $\hat n$, depends on the target extension length.

\noindent\textbf{Finding 3}: \textit{Non-uniform positional interpolation effectively extends LLM context window in both fine-tuning and non-fine-tuning settings.} 

While we've shown that our searched non-uniform position interpolation significantly improves the extension performance at 8k and 16k without fine-tuning, longer extensions require fine-tuning. As such, we fine-tune LLaMA2-7B with our searched RoPE for a 64k context window size (see Appendix for settings). As Table~\ref{tbl:finetune} shows, our method significantly outperforms PI and YaRN, both before and after fine-tuning LLaMA2-7B. This is due to our effective use of non-uniform positional interpolation, minimizing information loss and providing a better initialization for fine-tuning.

\noindent\textbf{Summary}. Our study uncovers two non-uniformities: varying RoPE dimensions and token positions. Utilizing these non-uniformities effectively in positional interpolation greatly improves LLM context extension performance.

	\vspace{-2ex}
\section{{\sysname}}
\vspace{-1ex}
\label{sec:method}
Motivated by the findings, we present {\sysname}, which first introduces an efficient search algorithm to fully exploit the two non-uniformities, and then uses it to extend LLM context window beyond 2 million tokens.

\vspace{-1ex}
\subsection{Problem Formulation}
\vspace{-1ex}
The two non-uniformities can lead to a vast solution space and introduce complexities in optimization. To address it,  we frame the multidimensional non-uniform position interpolation optimization problem as a  search problem.

For a LLM targeting a  context window size of $L'$ and lengthy input documents $\mathbf{X}$, where each $\mathbf{x}\in\mathbf{X}$ surpasses $L'$ in token length, 
we denote the original rotary angle of the $i^{th}$ dimension in RoPE embedding at token position $n$ as  $\frac{n}{\beta^i}$. The optimization problem is then formulated as follows:
\vspace{-2ex}
\begin{equation}
\begin{gathered}
		\label{eq:problem}
	%	\small
\underset {\mathbf{x}\in\mathbf{X}; \, |\mathbf{x}|\ge L'}{ \operatorname {arg\,min} } \,  \mathcal{L}\, \left( \text{LLM} (\text{RoPE}, \mathbf{X})\right),	\text{where \;} 
\\
\scalebox{0.93}{
$\underset {\begin{subarray}{l} i=0,\cdot\cdot,\frac{d}{2}-1;\\ n\in[ 0,|\mathbf{x}|);\end{subarray}}{\text{RoPE}_(n)}= {\left[\cdot\cdot,cos\left(\mathbb{I}(\hat{\lambda}_{i}, \hat{n})\times\frac{n}{\beta^i}\right),  sin\left(\mathbb{I}(\hat{\lambda}_{i}, \hat{n})\times\frac{n}{\beta^i}\right),\cdot\cdot\right] }$}\\
	\text{where \;} \mathbb{I}(\hat{\lambda}_{i},\hat{n})=\begin{cases}
	1&\mbox{\;  $n<\hat{n}$}\\
{\frac{1}{\lambda}_{i}}&\mbox{\; $n\geq\hat{n}$}
\end{cases}
	\end{gathered}
\end{equation}
where we introduce a set of rescale factors, $\mathbb{I}(\hat{\lambda}_{i},\hat{n})$,  to cover the two forms of non-uniformities.  $\hat{\lambda_i}$  and $\hat{n}$ denote the non-uniformity of RoPE dimensions and token positions, respectively. Specifically,
we use $\mathbb{I}(\hat{\lambda}_{i},\hat{n})$  to rescale the rotation angle for the $i^{th}$ RoPE dimension, where $\hat\lambda_{i}$ is the rescale factor and $\hat{n}$ is token position threshold. For initial $\hat{n}$-1 token positions, the rescale factor $\hat\lambda_{i}$ will not take effect, and the original RoPE rotary angle $\frac{n}{\beta^i}$ is used. For tokens at positions $n\ge\hat{n}$, the rescale factor is applied.

Given a target context window size of $L'$, our objective is to find the optimal rescale factors ($\mathbb{I}(\hat{\lambda}_{0},\hat{n})$, $\mathbb{I}(\hat{\lambda}_{1},\hat{n})$ ,...$\mathbb{I}(\hat{\lambda}_{i},\hat{n})$...) from the $1^{st}$ to $d^{th}$ RoPE dimension. As a result, the target $\text{LLM}$, with the rescaled $\text{RoPE}$, can achieve a minimum next token prediction loss, $\mathcal{L}$ (i.e., the perplexity), for input samples $\mathbf{X}$ with a token length of $L'$.

\vspace{-2ex}
\subsection{Searching the Non-uniform Position Interpolation}
\vspace{-1ex}
To solve the problem in Eq.~\ref{eq:problem}, we now introduce our simple yet highly effective method, which searches for the optimal RoPE rescale factors to fully exploit the multidimensional non-uniformities in position embedding.

\begin{table}[t]
	\caption{Search space for RoPE rescale factors. Tuples of three values represent the lowest value, highest, and step size.}
	\label{tbl:searchspace}
	\small
	\centering 
	\resizebox{0.48\textwidth}{!}{	\begin{tabular}{@{\hskip0pt}c@{\hskip2pt}|@{\hskip2pt}c@{\hskip2pt}|@{\hskip2pt}c@{\hskip0pt}}
			\toprule
			Non-uniformity&Notation&Search Space\\
			\hline
			RoPE dimension&$\lambda_i$& (1.0, extension ratio $s\times$1.25, 0.01)\\
			\midrule
			Starting tokens& $\hat{n}$ & \{0, 1, 2, 4, 8, 12, 16, 20, 24, 28, 32, 64, 128, 256\}\\		
			\hline 
		\end{tabular}}
	\end{table}
	
\noindent\textbf{Search space}. We  design a large search space to include the two non-uniformies.  Table~\ref{tbl:searchspace} illustrates the search space.  Specifically, we allow the search of a specialized rescale factor for each dimension in RoPE embedding. 
 To simply search space design, we search $\lambda_i$ and $\hat{n}$ instead of searching for $\mathbb{I}(\hat{\lambda}_{i},\hat{n})$, where  $\hat{\lambda}_{i}=1/\lambda_i$. 
 As shown in Table~\ref{tbl:searchspace}, $\lambda_i$ is allowed to search from a minimum value of 1.0 (i.e., direct extrapolation) to a maximum value of $s\times1.25$ (i.e., larger interpolation than PI) with a step size of 0.01, where $s$ is the target context window extension ratio.

$\hat{n}$ controls the number of initial token positions that are retained without position interpolation (i.e., use the original RoPE embedding). Empirically, we 
 allow $\hat{n}$ to search from \{0, 1, 2, 4, 8, 12, 16, 20, 24, 28, 32, 64, 128, 256\}. When $\hat{n}=0$, all token positions use the searched rescale factors.

\noindent\textbf{Evolution-based search}. Our search space in Table~\ref{tbl:searchspace} spans numerous positional interpolation solutions, posing a significant challenge for efficient exploration.  For example, a $s=4\times$ extension leads to  $400^{128/2}\times14$=4$\times10^{167}$ choices. With larger extension ratio,  the search space  expands exponentially.  
To address this, we use evolution search~\cite{guo2020single}  and introduce  two optimization techniques to greatly boost search efficiency. Algorithm~\ref{alg:search} illustrates the overall search procedure.

\noindent\textit{Optimized initial population generation}. Instead of initializing a population  of $P$ rescale factors randomly, we add the three RoPE rescale factors corresponding to PI, NTK, and YaRN as individuals into the initial population. For the remaining $P$-3 individuals, we randomly mutate the three rescale factors with a probability of $p$.

\noindent\textit{Monotonically non-decreasing constraint}. 
After generating the initial population, we compute LLM perplexity for each individual. Specifically, we apply the corresponding RoPE rescale factors to the target LLM and compute the perplexity of  input $\mathbf{X}$. The top-k individuals become parents for evolution. However, the vast search space can cause naive mutation and crossover to explore poor solutions, leading to unnecessary perplexity computations. This is particularly inefficient when $L'$ is large, given the time-consuming inference  of each perplexity calculation.

To address this, we impose a non-decreasing monotonicity constraint on the sampled RoPE rescaled factors: $\lambda_i\le \lambda_{i+1}$.
Only RoPE that satisfies this constraint is applied to LLM for perplexity evaluation, significantly reducing the search costs. Specifically,   we require that $\lambda_i$ increases monotonically with the RoPE dimension (i.e., $i$=0,...,63). This dimensional monotonicity is based on the NTK theory~\cite{jacot2018neural,tancik2020fourier,ntk}, suggesting that lower dimensions with higher frequency requires less interpolation (i.e., a smaller $\lambda_i$), and higher dimensions with lower frequency can do more interpolation (i.e.,  a larger $\lambda_i$).

\begin{algorithm}[t]
	\small
	\caption{The search algorithm}
	\textbf{Input:} target LLM, input samples $\mathbf{X}$,   population size $P$, mutation size $N_1$, crossover size $N_2$, max iterations $\mathcal{T}$, mutate probability $p$\\
	\vspace{-2.5ex}

	\begin{algorithmic}[1]
		\label{alg:search}
		\STATE Top-k=$\phi$;	
		\STATE $\text{P}_0$=\textit{Initialize\_population\_with\_optimization} ($P$, $\mathbf{X}$, $p$); 
		\FOR{$i$=$1$ to $\mathcal{T}$}
		\STATE \textit{Compute\_perplexity} (LLM, $\text{P}_{i-1}$, $\mathbf{X}$);
		\STATE  Top-k = \textit{Update\_Topk} (Top-k, $\text{P}_{i-1}$);
		\STATE$\text{P}_{mutation}$=\textit{Mutation\_with\_mono\_constraint} (Top-k, $N_1$, $p$); 
		\STATE $\text{P}_{crossover}$=\textit{Crossover\_with\_mono\_constraint} (Top-k, $N_2$);
		\STATE $\text{P}_i$ = $\text{P}_{mutation}$ $\cup$ $\text{P}_{crossover}$ $\cup$ Top-k;
		\ENDFOR
		\STATE Return the individual with lowest perplexity in Top-k;
	\end{algorithmic}
\end{algorithm}

\begin{figure}[t]
	\centering
	\includegraphics[width=1\columnwidth]{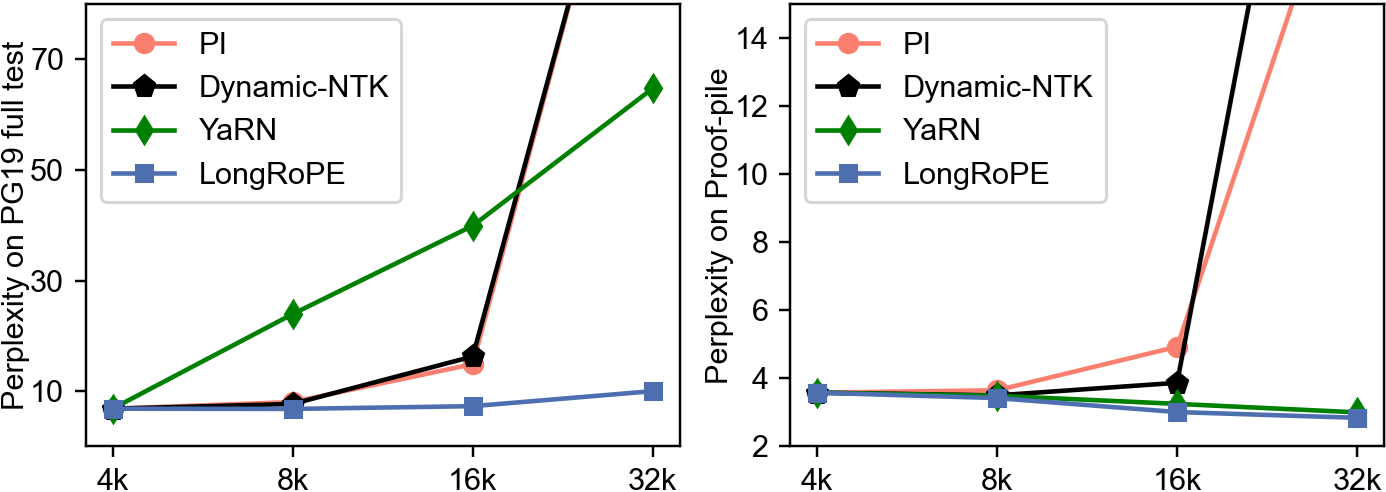}
	\vspace{-5ex}
	\caption{LLaMA2-7B perplexity on PG19 and Proof-Pile after extension using different methods, measured without fine-tuning. By fully exploiting the non-uniformities, {\sysname} achieves an \textbf{8$\times$ extension without fine-tuning}.}
	\label{fig:non-ft}
\end{figure}

\noindent\textbf{8$\times$ extension without fine-tuning}. Our evolutionary search  effectively identifies non-uniform RoPE rescale factors, preserving key dimensions and positions to minimize interpolation-induced information loss. As depicted in Fig.\ref{fig:non-ft}, our method is able to extend LLaMA2's context window from 4k to 32k without fine-tuning. In contrast, existing methods such as PI, and non-uniform NTK and YaRN cause perplexity to spike after 2$\times$ extension.

\vspace{-1.5ex}
\subsection{Extending LLM Context Window to 2048K}
\label{sec:extendto2048k}
\vspace{-1ex}
\noindent\textbf{Progressive extension to 2048k}. 
We now introduce our method  to extend the context window of pre-trained LLMs from the traditional 4k to over 2048k. As demonstrated, our non-uniform positional interpolation can achieve 8$\times$ extension without fine-tuning. 
For larger extensions (i.e., 512$\times$) is required, fine-tuning is necessary. One method is to search for RoPE rescaled factors under the target 2048k size and then fine-tune. 
However, this faces challenges due to the prohibitively expensive training resources. 
Moreover,  based on our experiences, it's challenging to well fine-tune the LLMs under a  large extension ratio (see Appendix). 

Fortunately, {\sysname} is effective for both the original and fine-tuned extended LLM. Therefore, we introduce an efficient, progressive method that achieves the target 2048k with just 1k fine-tuning steps at within 256k training length.

\noindent$\Diamond$ \textit{Extending pre-trained LLM to 256k with {\sysname} search}. Taking LLaMA2 as an example, we conduct search for target context window size of 128k and 256k. The extension ratio at this stage is 32$\times$ and 64$\times$, respectively.

\noindent $\Diamond$ \textit{Fine-tuning to 256k}. Then, we fine-tune the pre-trained LLM to achieve the context window size of 256k. 	Specifically, we first fine-tune LLaMA2 for 400 steps using the RoPE rescaled factors for 128k. Then, we replace the RoPE rescaled factors to 256k on the finished checkpoint and conduct an additional 600 steps of fine-tuning. This method proves more efficient than directly fine-tuning to 256k.

\noindent $\Diamond$ \textit{Extending fine-tuned extended LLM to 2048k with {\sysname} search}. Finally, we perform a secondary  search on the fine-tuned 256k-length LLM. This ultimately results in an extremely large context window size of 2048k without further fine-tuning. The final extension ratio is  $512\times$.	

\begin{table*}[ht]
	\centering
	\fontsize{8}{8} \selectfont
	\caption{Proof-pile perplexity of models with various positional interpolation methods.  ft: the context window size used in fine-tuning. Even with a context window 16$\times$ longer than current long-context models, our models also outperform them within 256k context length.}
	\label{tbl:proof-pile}
	\begin{tabular}
		{ccccccccccc}
		\toprule
		Base& Model& Context &  Extension& \multicolumn{7}{c}{Evaluation Context Length}\\
		LLM		&Name&Window&Method&4096&8192&32768&65536&98304&131072&262144\\
		\hline
		\multirow{8}{*}{LLaMA2-7B}& LLaMA2-7B& 4k&- &\bf 3.58 &$>$10$^4$&$>$10$^4$&$>$10$^4$&$>$10$^4$&$>$10$^4$&$>$10$^4$\\
		& Together&32k&PI&3.69& 3.50&2.64&$>$10$^2$&$>$10$^3$&$>$10$^4$&$>$10$^4$\\
		&LongLoRA&100k&PI&3.83& 3.62 &2.68 & 2.44& 2.33& 9.89&$>$10$^3$\\
		&Code LLaMA& 100k& NTK &3.95&3.71&2.74&2.55&2.54&2.71&49.33\\
		&YaRN ($s$=16)&64k&YaRN&3.69& 3.51&2.65&2.42&$>$10$^1$&$>$10$^1$&$>$10$^4$\\
		&YaRN ($s$=32)&128k&YaRN&3.75&3.56&2.70&2.45&2.36&2.37&99.64\\
		&\textbf{{\sysname}-2048k} (ft=128k) &\bf 2048k&{\sysname} &3.71&\bf 3.50 & \bf 2.60& \bf 2.36&\bf2.27  &\bf 2.26 &\bf 1.88\\
		&\textbf{{\sysname}-2048k} (ft=256k) &\bf 2048k&{\sysname} &3.85&3.65 &\bf 2.63 & \bf 2.38& \bf 2.28& \bf 2.26&\bf 1.87\\
		\hline
		\multirow{5}{*}{Mistral-7B}&{Mistral v0.1}&8k&-&\bf 3.09&2.96&$>$10$^2$&$>$10$^3$&$>$10$^3$ & $>$10$^3$&$>$10$^4$\\
		&YaRN ($s$=8) &64k& YaRN&3.18& 3.04&2.37 & 2.20& 10.39&57.4&$>$10$^4$\\
		&YaRN ($s$=16) & 128k & YaRN &3.21& 3.08&2.41 & 2.24&2.18 & 2.19 &4.91\\
		&\textbf{{\sysname}-2048k} (ft=128k)&\bf2048k&{\sysname} &3.20&\bf 3.04&\bf 2.36 &\bf 2.18 &\bf 2.13 &\bf 2.13 &\bf 1.85\\
		&\textbf{{\sysname}-2048k} (ft=256k)  &\bf2048k&{\sysname} &3.20&\bf 3.04 &\bf 2.36 &\bf 2.18 &\bf 2.13 &\bf 2.14 &\bf 1.84\\
		\hline
	\end{tabular}
\end{table*}

\begin{table*}[t]
	\centering
	\fontsize{8}{8} \selectfont
	%	\small
	\caption{Perplexity evaluation on Books3 dataset.  Without additional fine-tuning, our {\sysname}-2048k models, with a training context window size of 128k and 256k, effectively scale to an extremely long context size of 2048k. 1k=1024 tokens.}
	\label{tbl:books3}
	\resizebox{1\textwidth}{!}{	\begin{tabular}
			{@{\hskip0pt}c@{\hskip4pt}c@{\hskip4pt}c@{\hskip4pt}cccccccccc@{\hskip0pt}}
			\toprule
			Base& Model& Context &  Extension& \multicolumn{9}{c}{Evaluation Context Length}\\
			LLM	&Name&Window&Method&8k&16k&32k&64k&128k&256k&512k&1024k&2048k\\
			\hline
			\multirow{6}{*}{LLaMA2-7B}& 
			LongLoRA&100k&PI&6.99&6.80&6.66&6.59 &20.57 & 246.45&$>$10$^3$ & $>$10$^4$ &$>$10$^4$\\
			&Code LLaMA& 100k& NTK &7.68&7.49&7.38&7.88&9.80&98.30&$>$10$^3$&$>$10$^4$&$>$10$^4$\\
			&YaRN ($s$=16)&64k&YaRN&\bf 6.33&\bf6.20& \bf6.11&\bf 6.06&$>$10$^4$&$>$10$^4$&$>$10$^4$&$>$10$^4$&$>$10$^4$\\
			&YaRN ($s$=32)&128k&YaRN&6.38&6.25&6.16&6.11&\bf 6.12&$>10^4$&$>$10$^4$&$>$10$^4$&$>$10$^4$\\
			&\textbf{{\sysname}-2048k} (ft=128k) &\bf2048k&{\sysname} &6.55&6.35& 6.24& 6.18&6.17&\bf 6.17&\bf 6.36 &\bf 6.83 &\bf 7.80\\
			&\textbf{{\sysname}-2048k} (ft=256k)  &\bf2048k&{\sysname} &6.81&6.66& 6.31&6.27 &6.21&\bf 6.17&\bf 6.17 & \bf 6.35&\bf 7.08\\
			\hline 
			\multirow{5}{*}{Mistral-7B}&{Mistral v0.1}&8k&-&\bf6.32&66.61&$>$10$^2$&$>$10$^3$&$>$10$^3$&$>$10$^3$&-&-&-\\
			&YaRN ($s$=16) & 64k& YaRN&6.59&6.48& 6.42&6.45&104.15 & 727.20&$>10^3$ &$>10^4$&$>10^4$\\
			&YaRN ($s$=32) & 128k & YaRN &6.70& 6.63&6.65&6.72&6.85 & 99.90& $>10^3$&$>10^4$& $>10^4$\\
			&\textbf{{\sysname}-2048k} (ft=128k)  &\bf2048k&{\sysname} & 6.64&\bf 6.48&\bf 6.39&\bf 6.45& \bf 6.64& \bf 7.08&\bf 7.71 & \bf 8.93&\bf 12.78\\
			&\textbf{{\sysname}-2048k} (ft=256k)  &\bf2048k&{\sysname} & 6.63&\bf 6.48&\bf 6.38&\bf 6.43&\bf 6.68 &\bf 7.15 & \bf 7.98&\bf 9.42&\bf 13.71\\
			\hline	\end{tabular}}
\end{table*}

\begin{table}
	\caption{Perplexity evaluation within 256k context length on PG19.}
	\small
	\label{tbl:pg19}
	\resizebox{0.48\textwidth}{!}{	\begin{tabular}{@{\hskip0pt}c@{\hskip4pt}c@{\hskip4pt}c@{\hskip1pt}ccc@{\hskip2pt}c@{\hskip0pt}}
			\toprule
			Base& Model&Context&Extension & \multicolumn{3}{@{\hskip2pt}c@{\hskip0pt}}{Evaluation Context Length}\\
			LLM&Name&Window&Method&8k&64k&128k\\
			\midrule
			\multirow{4}{*}{LLaMA2-7B}	&LongLoRA&100k& PI& 7.16&6.81&$>10^3$\\
			&Code LLaMA & 100k& NTK&7.58&8.92& 16.80\\
			&{\sysname}-2048k (ft=128k) & 2048k & {\sysname}&\bf 6.98&\bf 6.59&\bf 6.35 \\
			&{\sysname}-2048k (ft=256k)& 2048k &{\sysname} &7.37&\bf 6.64&\bf 6.31\\
			\hline
			\multirow{5}{*}{Mistral-7B}	
			&YaRN-64k&64k& YaRN&7.12 &7.17&$>10^3$\\
			&YaRN-128k & 128k& YaRN&7.30&7.53&7.32 \\
			&{\sysname}-2048k (ft=128k) & 2048k & {\sysname}&7.13 &\bf 7.01&\bf 7.02\\
			&{\sysname}-2048k (ft=256k)& 2048k &{\sysname} &\bf 7.10&\bf 6.98&\bf 7.13\\
			\hline
	\end{tabular}}
\end{table}
\noindent\textbf{Shorter context window recovery}.  After extending to an extremely long 2048k context window, we notice a performance drop within the original context window. This is a known issue of positional interpolation~\cite{pi}, as it forces position embedding in higher dimensions within the original context window to reside in a much narrower region, negatively affecting the language model's performance. With a 512$\times$ extension ratio, positions within the original 4k context window become particularly crowded.

To mitigate this, we perform an extra evolution search on the extended LLM to adjust RoPE rescale factors for short context lengths (e.g., 4k and 8k). 
We reduce the maximum allowed searched $\lambda$ due to less positional interpolation required for shorter lengths. During inference, the LLM dynamically adjusts the corresponding RoPE rescale factors.

\vspace{-2ex} 
 \section{Experiments}
 \vspace{-1ex} 
 \label{sec:exp}
\subsection{Setup}
\noindent\textbf{Evaluation Tasks and models}. We apply {\sysname} on LLaMA2-7B and Mistral-7B, and evaluate the performance on three aspects: (1) perplexity of extended-context LLMs on long documents; (2) Passkey retrieval task that measures a model's ability to retrieve a simple passkey from a sea of irrelevant text; and (3) Standard LLM benchmarks within a short 4096 context window size.

\noindent\textbf{Fine-tuning}. For LLaMA2, we use a learning rate of 2e-5 with linear decay and a global batch size of 32. We fine-tune for 400 steps on  Redpajama~\cite{redpajama} dataset, chunked into  128k segments bookended with the BOS and EOS tokens. Then, based on the finished checkpoint, we train an additional 600 steps to achieve 256k context window. The 128k context size is trained on 8 A100 GPUs with the distributed training system~\cite{cube}, while the 256k requires 16 A100 GPUs. 
In the case of Mistral, a constant learning rate of 1e-6 and a global batch size of 64 are used. For both 128k and 256k  models, we follow the setting in YaRN~\cite{yarn}, with  400 steps on the Together Computer's Long-Data Collections~\cite{mistraldata} using 16k sequence length. We use 4 A100 GPUs for training.

\noindent\textbf{Search}. For target window size within 256k,
we use:  $P$=64, $N_1$=$N_2$=16, $p$=0.3, $\mathcal{T}$=40, and select top-32 for mutation/crossover in each iteration. Perplexity is calculated using 5 random PG19 validation set samples, with a minimum length requirement of the target context length. For windows over 512k, 
 we halve the population, mutation, and crossover sizes. Perplexity is measured on 3 random samples from Pile-Books3~\cite{pile} validation set. 

\noindent\textbf{Baselines}. To reach 2048k, we fine-tuned models with 128k and 256k context windows. This yields {\sysname}-2048k (ft=128k) and {\sysname}-2048k (ft=256k) for LLaMA2 and Mistral, respectively.
We compare the four models with state-of-the-art context window extension baselines, specifically open-sourced LLMs fine-tuned after positional interpolation using PI, NTK and YaRN.  This includes Together-32k~\cite{together}, Code LLaMA~\cite{codellama}, LongLoRA-full-FT-100k~\cite{longlora}, YaRN-LLaMA and YaRN-Mistral~\cite{yarn}.

% \vspace{-1ex} 
\subsection{Main Results}
%\vspace{-1ex} 
\label{sec:mainresult}
\noindent\textbf{Long sequence language modeling within 256k}. 
We begin by comparing  with state-of-the-art extended LLMs within a 256k evaluation length. We use two datasets to demonstrate our generalizability: Proof-pile~\cite{pg19} and PG19~\cite{pile} test splits. We evaluate perplexity at various context lengths using  sliding window of 256. For PG19, we use the whole test split of 100 documents. For Proof-pile, we follow YaRN~\cite{yarn} to randomly select 10 samples, each with at  least 128k lengths. 

Table~\ref{tbl:proof-pile} and Table~\ref{tbl:pg19} compare the perplexity of LLaMA2 and Mistral extended via different interpolation methods on Proof-pile and PG19, respectively. We highlight two key observations: \textbf{(1)} our extended models show an overall decreasing perplexity trend   from 4k to 256k evaluation lengths, proving their abilities to leverage longer context. \textbf{(2)} Even with a context window 16$\times$ longer, a condition typically challenging for maintaining performance at shorter lengths, our {\sysname}-2048k models outperform state-of-the-art baselines within 256k context length.

\noindent\textbf{Long sequence language modeling beyond 2000k}. 
 To evaluate the effectiveness on extremely long documents, we use the Books3~\cite{pile} dataset. For evaluation efficiency, we randomly select 20 books, each  exceeding 2048k in length, and use a sliding window of 256k. 

As shown in Table~\ref{tbl:books3}, {\sysname} successfully extends LLaMA2-7B and Mistral-7B's context window to 2048k, while also achieving perplexity comparable or superior to baselines within shorter lengths of 8k-128k. We also observe notable performance differences between the 2048k LLaMA2 and Mistral.  Mistral outperforms baselines at shorter lengths, but perplexity exceeds 7 beyond 256k. LLaMA2 performance aligns with expectations: the perplexity decreases gratefully with longer contexts, with marginal increases at 1024k and 2048k. 
Moreover, on LLaMA2, {\sysname}-2048k performs better at a fine-tuning length of 256k over 128k, due to the smaller secondary extension ratio (i.e., 8$\times$ vs. 16$\times$). In contrast, Mistral performs better at fine-tuning window size of 128k.  The main reason is that for Mistral's 128k and 256k fine-tuning, we follow YaRN's setting to use a 16k training length, which affects Mistral's ability to further extend context window after fine-tuning.

\noindent\textbf{Passkey retrieval}. We now study the effective context window size in generation tasks. We follow a synthetic evaluation task of passkey retrieval proposed by ~\cite{passkey}. In this task, the model is asked to retrieve a random passkey (i.e., a five-digit number) hidden in long document. The prompt template is detailed in appendix. 
We perform 10 iterations of the passkey retrieval task with the passkey placed at a random location uniformly distributed across the evaluation context length.

\begin{figure}[t]
	\centering
	\includegraphics[width=1\columnwidth]{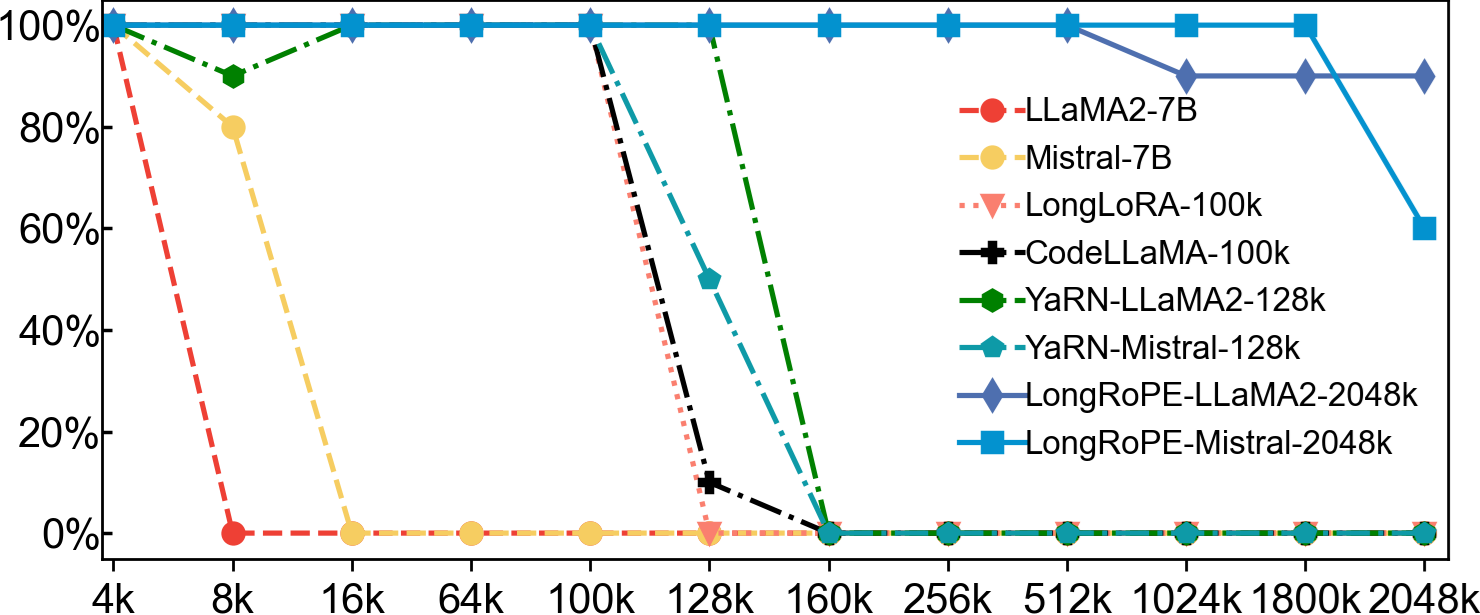}
	\vspace{-5ex}
	\caption{Passkey retrieval accuracy of long-context LLMs. It showcases the remarkable ability of our models to accurately retrieve a passkey from a vast pool of million-level tokens.}
	\label{fig:passkey}
\end{figure}

 Fig.~\ref{fig:passkey} shows the retrieval accuracy comparison  with baselines. Existing models' accuracy rapidly drops to 0 beyond 128k.  In contrast, despite the very challenging task of retrieving a passkey from million-level tokens, our {\sysname}-LLaMA2-2048k (ft=256k)  manage to maintain a high retrieval accuracy ($\ge$90\%) from 4k to 2048k. {\sysname}-Mistral-2048k (ft=128k) keeps 100\% accuracy up to 1800k, dropping to 60\% at 2048k, aligning with expectations from Table~\ref{tbl:books3}, where the perplexity slightly increases at 2048k.

\noindent\textbf{Standard benchmarks within original context window}. We evaluate {\sysname}-2048k models on the original context window using 
Hugging Face Open LLM Leaderboard~\cite{huggingfaceboard} in zero-shot and few-shot settings. We use 25-shot ARC-Challenge~\cite{allenai:arc}. 10-shot HellaSwag~\cite{zellers2019hellaswag}, 5-shot MMLU~\cite{mmlu}, and 0-shot TruthfulQA~\cite{lin2021truthfulqa}. 

As Table~\ref{tbl:commonsense} shows, our models achieve comparable results on the original benchmark designed for a smaller context window, and even outperform the original Mistral on TruthfulQA by +0.5\%. {\sysname}-LLaMA2-2048k, fine-tuned at 256k, shows slightly more performance degradation, but remains within reasonable ranges for most tasks.

\begin{table}[t]
	\centering
	\small
	\caption{Comparison of long-context LLMs with original LLaMA2 and Mistral on the Hugging Face Open LLM benchmark. }
		\label{tbl:commonsense}
	\resizebox{0.48\textwidth}{!}{	\begin{tabular}{@{\hskip0pt}c@{\hskip3pt}c@{\hskip5pt}c@{\hskip5pt}c@{\hskip5pt}c@{\hskip5pt}c@{\hskip0pt}}
		\toprule
		\multicolumn{6}{c}{\bf(a) LLaMA2-7B with extended context window}\\
		\hline
\multirow{2}{*}{Model}&Context& \multirow{2}{*}{ARC-c}&\multirow{2}{*}{HellaSwag}&\multirow{2}{*}{MMLU}&\multirow{2}{*}{TruthfulQA}\\
&Window&&&&\\
	\midrule 
	Original LLaMA2-7B& 4k& 53.1&78.6&46.6&39.0\\
	\hdashline
 Together&32k&47.6&76.1&43.3&\bf 39.2\\
	 Code LLaMA &100k& 42.4&64.8&40.1& 37.1\\
	 YaRN ($s$=16) & 64k& 52.4& \bf 78.7& 42.4& 38.2\\
 YaRN ($s$=32) &128k& 52.2& 78.5& 41.8& 37.4\\
	{\sysname}-2048k (ft=128k)&2048k&\bf 52.9&76.5&\bf 43.4&38.8\\
	{\sysname}-2048k (ft=256k)&2048k&51.0&75.3&39.6&37.3\\
	\midrule
	\multicolumn{6}{c}{\bf(b) Mistral-7B with extended context window}\\\hline
Original Mistral-7B& 8k& 60.6&83.2 &63.6 &42.6 \\
		\hdashline
	MistralLite~\cite{mistrallite} & 16k & 59.2&\bf 81.6 &50.4 &38.3 \\
	 YaRN ($s$=16) & 64k&\bf59.3 &81.3 &61.3 &42.5 \\
	YaRN ($s$=32) &128k& 59.0&80.5 & 60.5&42.5 \\
	{\sysname}-2048k (ft=128k)&2048k& 59.0&81.2 & \bf 61.3&\bf 43.1 \\
	{\sysname}-2048k (ft=256k)
	 &2048k&59.2 & 80.9& 61.1&42.2\\
	\hline
	\end{tabular}}
\end{table}

\vspace{-1ex}
\subsection{Ablation Results}
\vspace{-1ex}

\begin{table}[t]
	\small 
	\centering
	\caption{Books3 perplexity comparison of extending LLaMA2-256k via different secondary positional interpolation methods.}
	\label{tbl:secondsearch}
	\resizebox{0.48\textwidth}{!}{	\begin{tabular}{@{\hskip0pt}c@{\hskip4pt}|cccc@{\hskip0pt}}
		\toprule
		Model&Extension& \multicolumn{3}{c}{Context Window Size}\\
		Name&Method&512k&1024k&2048k\\
		\midrule 
\multirow{3}{*}{LLaMA2-7B (ft=256k)}	&PI &6.60 &8.73 &20.17\\
	&	YaRN &6.39 &6.79 &8.27\\
	&\bf	{\sysname}   &\bf 6.17 &\bf 6.35 &\bf 7.08 \\
		\hline
	\end{tabular}}
\end{table}

\begin{table}[t]
	\small 
	\centering
	\caption{Ablation study on {\sysname} readjustment for performance recovery at shorter context lengths.}
	\label{tbl:shortperformance}
	\resizebox{0.48\textwidth}{!}{\begin{tabular}{@{\hskip0pt}c@{\hskip1pt}|c@{\hskip2pt}|cc@{\hskip0pt}|@{\hskip4pt}c@{\hskip0pt}}
			\toprule
			\multirow{2}{*}{FT Model}&With&\multicolumn{2}{@{\hskip4pt}c@{\hskip4pt}|@{\hskip0pt}}{Proof-Pile Perplexity}&LLM Benchmark\\
			&Recovery&4k &8k& Avg. Accuracy\\
			\hline 
			\multirow{2}{*}{LLaMA2-7B-2048k (ft=128k)}&$\times$  & 4.16&3.72 &49.3 \\
			&\checkmark &\bf 3.71 &\bf 3.50 &\bf 52.9\\
			\hline
			\multirow{2}{*}{LLaMA2-7B-2048k (ft=256k)}& $\times$ & 4.51&3.82 &47.9\\
			& \checkmark&\bf 3.85 &\bf 3.65&\bf 50.8\\
			\hline
			
	\end{tabular}}
\end{table}

\begin{table}[t]
	\small 
	\centering
	\caption{Ablation study on the two forms of non-uniformities.}
	\label{tbl:searchdimension}
	\resizebox{0.48\textwidth}{!}{	\begin{tabular}{@{\hskip0pt}c@{\hskip2pt}|cc|@{\hskip2pt}c@{\hskip0pt}}
			\toprule
			&\multicolumn{2}{@{\hskip4pt}c|@{\hskip4pt}}{LLaMA2-7B}&LLaMA2-7B (ft=256k)\\
			Methods&\multicolumn{2}{@{\hskip4pt}c|@{\hskip4pt}}{PG19 Perplexity}&Books3 Perplexity\\
			\cline{2-4}
			&16k&32k&2048k\\
			\hline
			Linear interpolation (PI) &14.88 & 136.30& 20.17\\
			RoPE dim (Ours)&7.28 &13.00&7.08\\
			RoPE dim+Start tokens (Ours)&\bf 7.22&\bf 11.51&\bf 7.08\\
			\hline
	\end{tabular}}
\end{table}

\noindent\textbf{Effectiveness of the second positional interpolation}. In our progressive extension strategy, we use our search algorithm to conduct a second non-uniform positional interpolation on the fine-tuned extended LLMs. We validate its effectiveness by running experiments on our fine-tuned LLaMA2-256k model.  We extend it to 512k, 1024k, and 2048k using PI and YaRN. As Table~\ref{tbl:secondsearch} shows, our non-uniform positional interpolation sustains a consistent level of perplexity.  In contrast, the perplexity under PI and YaRN quickly increases with the extension ratio.

\noindent\textbf{Effectiveness of recovery at shorter context lengths.} To mitigate performance loss at shorter context lengths, we readjust the RoPE factors for {\sysname}-2048k via our search algorithm. Specifically, we decrease the maximum allowable scale factors for the search to encourage less interpolation at short 4k and 8k lengths. 
 Table~\ref{tbl:shortperformance} shows the perplexity comparison of {\sysname}-LLaMA2-2048k on Proof-pile at 4k and 8k lengths, along with the average LLM benchmark accuracy.  The results clearly demonstrate a significant performance improvement at short context lengths.

\noindent\textbf{Analysis on the two forms of non-uniformities}. Finally, we ablate on the two non-uniformities to see how each part contributes to the performance. We setup two experiments: (i) extending LLaMA2-7B to short 16k and 32k using different methods—PI, searching for RoPE dimension only, and searching for both non-uniformities; (ii) extending our fine-tuned 256k-length LLaMA2 to 2048k following the same procedure.  The perplexity is evaluated without fine-tuning. 
As Table~\ref{tbl:searchdimension} shows, non-uniformity in RoPE dimension significantly reduces perplexity compared to PI's linear interpolation. Non-uniformity in token position clearly improves performance at 16k and 32k lengths  but does not show the same impact at 2048k, possibly due to the extremely long length. Preserving only the initial tokens without interpolation becomes non-useful, and we leave this as future work.  
	\vspace{-4ex}
\section{Related Works}
\vspace{-1ex}
In addition to methods based on position interpolation, this section discusses related works of other approaches.

\noindent\textbf{Retrieval-based} approaches  use an external memory module to memorize long past context and retrieval modules for related documents fetching at inference~\cite{longllama,wang2023augmenting,borgeaud2022improving}. These designs typically need explicit modifications on the LLM architectures. Our work, in contrast, is more lightweight, with minor positional embedding modifications. We can also handle more long context tasks beyond retrieval, such as long document summarization and few-shot learning.

\noindent\textbf{Attention-based context window extensions}. Beyond positional embedding interpolation, some research achieves input context extension using the original LLM context window length by manipulating attention mechanisms~\cite{han2023lminfinite, streamingllm,parallelwindow}. The key idea is to mitigate the attention explosion issue caused by new positions using novel attention masks. 
These efforts and positional interpolation methods are  complementary.

\noindent\textbf{Fine-tuning based approaches} focus on how  to effectively fine-tune pre-trained LLMs with modified position embeddings for longer context. Works like Code LLaMA~\cite{codellama}, LLaMA2 Long~\cite{xiong2023effective} and ScaledRoPE~\cite{liu2023scaling} choose a very large base value for RoPE and fine-tune on the target length. Our method offers flexibility for various target lengths and can achieve beyond 2M length.  More recently, as fine-tuning for long context lengths (i.e., over 128k) demands substantial GPU resources,  LongLoRA~\cite{longlora} and PoSE~\cite{zhu2023pose} are proposed to mitigate this overhead. 
Our method is orthogonal to these efficient fine-tuning works.

	\vspace{-1ex}
\section{Conclusion}
\vspace{-1ex}
In this work, we present {\sysname}, a method that remarkably extends the context length of LLMs to an unprecedented 2048k, while maintaining their  capabilities within original shorter context window. We exploit two forms of non-uniformities in RoPE positional embedding using an efficient evolutionary search. This  offers twofold benefits: it provides good  initialization for fine-tuning and enables an 8$\times$ context window extension without fine-tuning. Building on this, we  propose a progressive extension strategy using 256k-length fine-tuned LLMs to reach a 2048k context window size without extra fine-tuning. Extensive experiments validate the effectiveness of {\sysname}. We envision that our {\sysname}-2048k models will enable many new long context applications and inspire further research.

\section*{Broader Impacts}
This paper presents work whose goal is to advance the field of Machine Learning. There are many potential societal consequences of our work, none which we feel must be specifically highlighted here.

	\balance
	\bibliography{ref}
	\bibliographystyle{icml2024}
	\newpage
	\appendix 
	\onecolumn
	\section{Appendix}

\subsection{Settings}
\noindent\textbf{Environments}. All our experiments are conduct on 16 A100 GPUs. We employ Flash Attention-2~\cite{dao2023flashattention2} to accelerate both training and inference. As the  GPU memory and computation time increase exponentially with the sequence length, it's challenging to serve the fine-tuning and inference with context length beyond 512k. %As a result, we utilize CUBE - an advanced version of ~\cite{cube}, to reduce both the training and inference costs. 
As a result, we utilize an internal platform, CUBE - an internal version of ~\cite{cube}, to reduce both the training and inference costs.

\noindent\textbf{Passkey prompt}. 
We follow existing literature~\cite{passkey, pi, yarn, longlora, zhu2023pose} for the document format of passkey retrieval.  We show the prompt template as follows:

\texttt{There is an important info hidden inside a lot of irrelevant text. Find it and memorize them. I will quiz you about the important information there.}

\texttt{The grass is green. The sky is blue. The sun is yellow. Here we go. There and back again. \underline{(repeat x times)}}

\texttt{The pass key is \textbf{17865}. Remember it. \textbf{17865} is the pass key.}

\texttt{The grass is green. The sky is blue. The sun is yellow. Here we go. There and back again. \underline{(repeat y times)}}

\texttt{What is the pass key? The pass key is}

The document length varies with the value of \texttt{x} and \texttt{y}. \texttt\textbf{17865} is the passkey number to retrieve. It is randomly sampled and varies at each testing time.

\subsection{Additional details on fine-tuning}

\begin{figure}[ht]
	\centering
	\includegraphics[width=0.9\columnwidth]{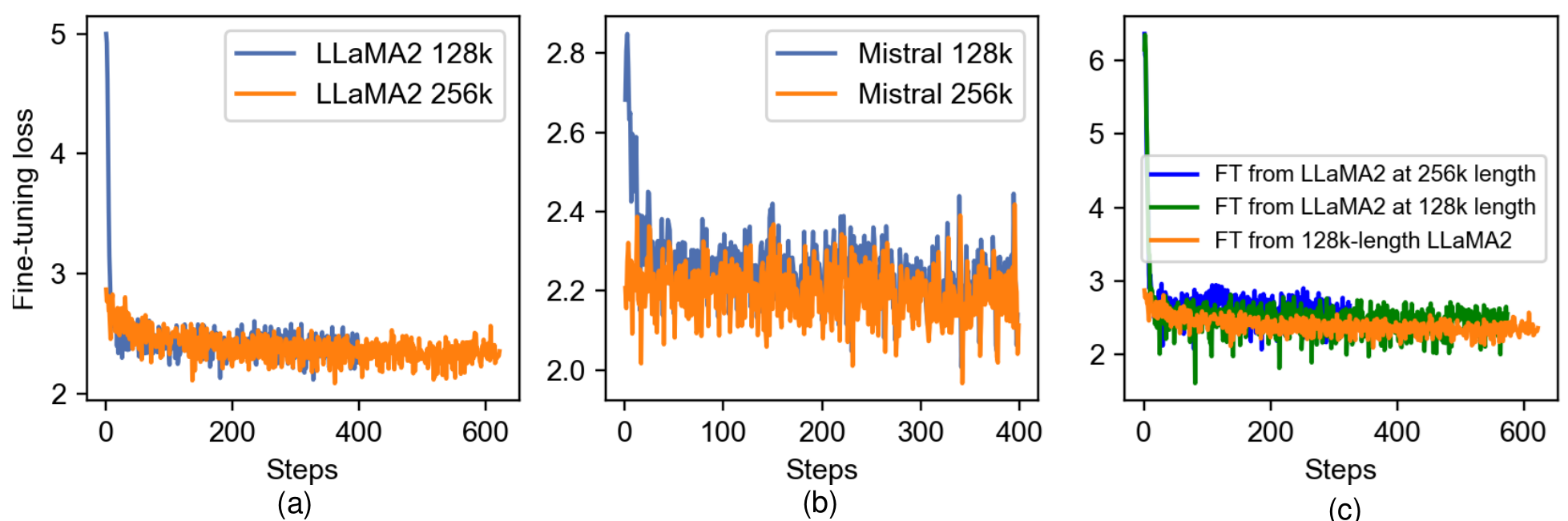}
	\vspace{-2ex}
	\caption{(ab): Loss curve in fine-tuning LLaMA2-7B and Mistral-7B with extended context window size. (c) The training loss of fine-tuning LLaMA2-7B with a 256k context window under different fine-tuning settings.}
	\label{fig:ftloss}
\end{figure}

As introduced in Section~\ref{sec:mainresult}, we fine-tune two context window lengths, namely 128k and 256k, for both LLaMA2 and Mistral. Specifically, the model with a 256k context window begins its fine-tuning from the 128k-length checkpoint. Fig.~\ref{fig:ftloss}(ab) illustrates the training loss for LLaMA2 and Mistral during this fine-tuning process. We highlight three key observations: \textbf{(1)} The model with a 128k context window experiences a large initial loss due to a 32$\times$ extension. However, the loss rapidly decreases after a few steps. \textbf{(2)}  LLaMA2 and Mistral employ different fine-tuning settings. Mistral achieves the desired long context window by fine-tuning on 16k-length data, while LLaMA2 necessitates text lengths that match the context window size. Furthermore, we adopt YaRN’s strategy of using a constant learning rate. As a result, it can be observed that Mistral’s loss begins to fluctuate after dropping to around 2.2. \textbf{(3)}  For both Mistral and LLaMA2, the model with a 256k context window, which starts fine-tuning from the 128k checkpoint, exhibits a low initial training loss. This suggests that fine-tuning from 128k-length checkpoints is effective and significantly facilitates convergence.

We also explore different settings to fine-tune LLaMA2 with 256k context window. As shown in Fig.~\ref{fig:ftloss}(c), we experiment with two additional settings: (i) using the RoPE rescale factors corresponding to 256k, we directly fine-tune on LLaMA2-7B, and (ii) using RoPE rescale factors for 256k, we fine-tune on LLaMA2-7B, but truncate the text lengths to 128k. The loss curves are displayed in Fig.~\ref{fig:ftloss}(c). 
We observe that using 128k text lengths to fine-tune a model with a 256k context window results in a sharp increase in the initial loss. Directly fine-tuning from LLaMA2-7B to achieve 256k results in a relatively slow decrease in loss. Table~\ref{tbl:ft} shows the test perplexity on Proof-Pile for checkpoints from three different settings. This indicates that our current approach of fine-tuning from a 128k-checkpoint is the most effective.

\begin{table}[h]
	\small 
	\centering 
	\caption{Proof-pile perplexity of extended LLaMA2-7B via different fine-tuning settings.  Tuples of three values represent the fine-tuning text length, context window size and initial checkpoint.}
	\label{tbl:ft}
		\begin{tabular}{@{\hskip0pt}c@{\hskip3pt}c@{\hskip3pt}c@{\hskip3pt}c@{\hskip3pt}c@{\hskip3pt}c@{\hskip0pt}}
			\toprule
			Method &\multicolumn{5}{c}{Evaluation Context Length}\\
			(fine-tune $L'$, $L'$, base LLM)&32768&65536&98304&131072&262144\\
			\midrule
			(128k, 256k, LLaMA2-7B)  & 9.75& 6.56&5.15 &5.19 &2.21\\
			(256k, 256k, LLaMA2-7B)  & 4.51&2.87 & 2.53& 2.39&1.95\\
			(128k, 256k, LLaMA2-7B (ft=128k)  &\bf 2.66 &\bf 2.38 &\bf2.28 &\bf2.26 &\bf 1.87 \\
			\hline
	\end{tabular}%}
\end{table}

\noindent\textbf{Fine-tuning cost}. LLaMA2-128k uses 8 A100 GPUs for a week to fine-tune 400 steps. LLaMA2-256k doubles the resources to 16 A100 GPUs for two weeks to fine-tune 600 steps. For Mistral-128k and 256k, with a training length of 16k, we employ 4 A100 GPUs for a 2-day fine-tuning period.

\subsection{Additional details on the search}

\begin{figure}[h]
	\centering
	\includegraphics[width=0.9\columnwidth]{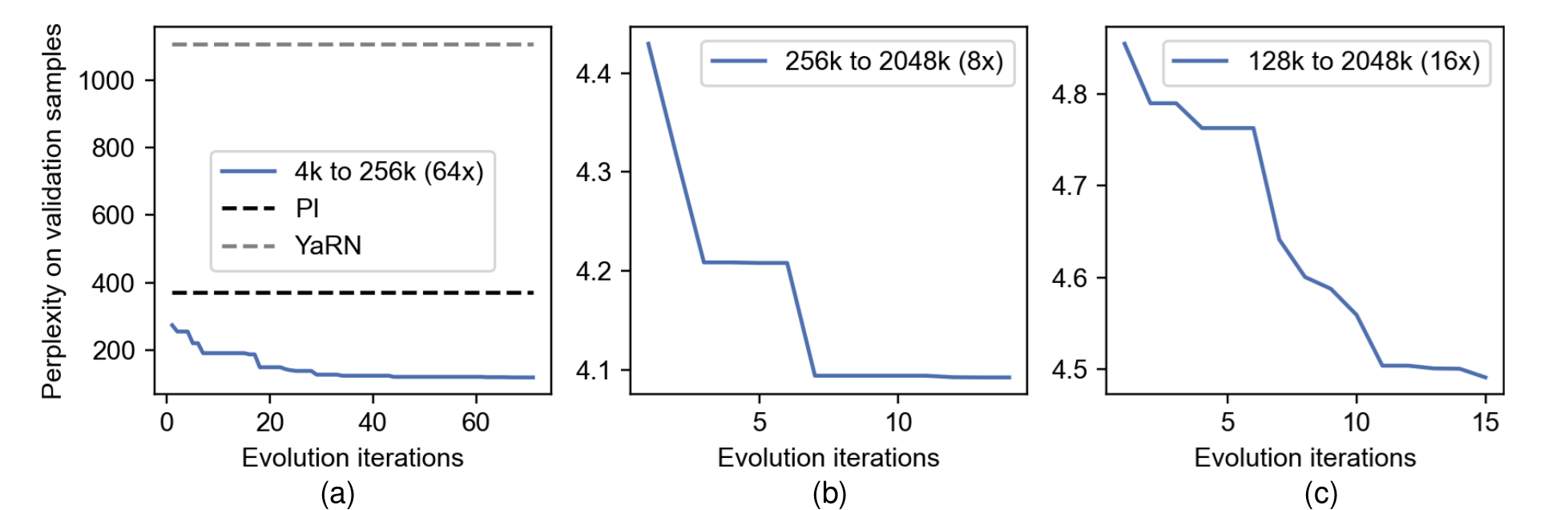}
	\vspace{-2ex}
	\caption{Perplexity on the validation samples at each evolution search iteration.  (a) The 64$\times$  extension for LLaMA2-7B to reach 256k context window size. (b) The 8$\times$ extension for LLaMA2-7B-256k to reach 2048k context window size. (c) The 16$\times$ extension for LLaMA2-7B-128k to reach 2048k context window size.    }
	\label{fig:searchefficiency}
\end{figure}

\noindent\textbf{Search efficiency}. Fig.~\ref{fig:searchefficiency} illustrates the perplexity on the validation samples at each evolution search iteration. We can see that our search algorithm can efficiently find high-quality non-uniform RoPE rescale factors. Specifically, on the 256k context window search (Fig.~\ref{fig:searchefficiency}(a)), after the first iteration, we can find solutions significantly better than PI and YaRN. As searching more iterations, we can significantly reduce the validation perplexity from 273.27 from 118.47.  Furthermore, we can observe that YaRN, as the previous state-of-the-art non-uniform interpolation method, performs even worse than PI (linear interpolation) at the 64$\times$ extension. This also indicates that human-heuristic-based non-uniform interpolation is challenging to perform well in all scenarios. 

For the extremely long context window at 2048k, we use the fine-tuned 128k and 256k context window's LLaMA2-7B for 16$\times$ and 8$\times$ extension, respectively. As shown in Fig.~\ref{fig:searchefficiency}(bc), as expected, the perplexity of the 16$\times$ extension is larger than that of the 8$\times$ extension. Additionally, due to the time required for a single perplexity evaluation at 2048k is about 50 minutes, the search iterations are constrained. If more search time is allowed, it's highly possible to search better results.  

\noindent\textbf{Search cost}. The search cost is primarily depending on the time required to evaluate the perplexity of input context at a given context window size. For context window lengths up to 256k, the total search time is relatively quick, achievable within 3 days using a single A100 GPU. For a 512k context window, we employ 2 A100 GPUs. For larger context windows of 1024k and 2048k, we utilize 4 and 8 A100 GPUs respectively, managing to keep the total search time within a 5-day limit.

\end{document}